\documentclass[pmlr]{jmlr}

\RequirePackage{graphicx}
 \usepackage{booktabs}
\usepackage{longtable}
\usepackage{float} 
\usepackage{lscape}
\usepackage{threeparttablex}

\makeatletter
\def\set@curr@file#1{\def\@curr@file{#1}} 
\makeatother
\usepackage[load-configurations=version-1]{siunitx} 

\theorembodyfont{\upshape}
\theoremheaderfont{\scshape}
\theorempostheader{:}
\theoremsep{\newline}

\jmlrvolume{252}
\jmlryear{2024}
\jmlrworkshop{Machine Learning for Healthcare}

\usepackage{todonotes}

\title[Early Prediction of Causes (not Effects) in Healthcare]{Early Prediction of Causes (not Effects) in Healthcare by Long-Term Clinical Time Series Forecasting}

 \author{\Name{Michael Staniek} \Email{staniek@cl.uni-heidelberg.de}\\
 \Name{Marius Fracarolli} \Email{fracarolli@cl.uni-heidelberg.de}\\
  \Name{Michael Hagmann} \Email{hagmann@cl.uni-heidelberg.de}\\
  \Name{Stefan Riezler} \Email{riezler@cl.uni-heidelberg.de}\\
  \addr Department of Computational Linguistics and\\ Interdisciplinary Center for Scientific Computing (IWR)\\ Heidelberg University, Germany}

\begin{document}

\maketitle

\begin{abstract}
Machine learning for early syndrome diagnosis aims to solve the intricate task of predicting a ground truth label that most often is the outcome (effect) of a medical consensus definition applied to observed clinical measurements (causes), given clinical measurements observed several hours before.  
Instead of focusing on the prediction of the future effect, we propose to directly predict the causes via time series forecasting (TSF) of clinical variables and determine the effect by applying the gold standard consensus definition to the forecasted values.
This method has the invaluable advantage of being straightforwardly  interpretable to clinical practitioners, and because model training does not rely on a particular label anymore, the forecasted data can be used to predict any consensus-based label. 
We exemplify our method by means of long-term TSF with Transformer models, with a focus on accurate prediction of sparse clinical variables involved in the SOFA-based Sepsis-3 definition and the new Simplified Acute Physiology Score (SAPS-II) definition.
Our experiments are conducted on two datasets and show that contrary to recent proposals which advocate set function encoders for time series and direct multi-step decoders, best results are achieved by a combination of standard dense encoders with iterative multi-step decoders.
The key for success of iterative multi-step decoding can be attributed to its ability to capture cross-variate dependencies and to a student forcing training strategy that teaches the model to rely on its own previous time step predictions for the next time step prediction.
\end{abstract}

\section{Introduction}
\label{sec:intro}

Early detection of syndromes like Sepsis 
is key to prevent a rapid progression to deadly stages by timely clinical invervention \citep{FerrerETAL:14,RuddETAL:20}. 
Machine learning from electronic health records (EHRs) bears the big promise of enabling early prediction of syndromes from historic measurements of vital and physiological parameters and laboratory test results.
Inspired by supervised machine learning,  early syndrome prediction is often framed as predicting a future label given clincial variables observed so far.
Crucial prerequisites in this regime are the availability of syndrome labels and the exact time of syndrome onset. However, such labels are often not routinely documented by clinicians. Furthermore, in the rare case where they are part of the information provided in EHRs, caution is advised in interpreting the chart time of diagnosis as the true time of syndrome onset.\footnote{For example, in the two datasets that to our awareness provide expert annotations of the exact time of sepsis diagnosis, one dataset only reports a normalized daily diagnosis time of 2 p.m. \citep{SchamoniETAL:19,LindnerETAL:22}, and the other \citep{PollardETAL:18} explicitly warns that the "diagnoses that were documented in the ICU stay by clinical staff [...] may or may not be consistent with diagnoses that were coded and used for professional billing or hospital reimbursement purposes." (Documentation of diagnosis field of eICU Collaborative Research Database, available at 
\url{https://eicu-crd.mit.edu/eicutables/diagnosis/}.)} Most machine learning approaches to early prediction of syndromes thus resort to automatically creating ground truth labels by applying consensus definitions, for example, the Sepsis-3 definition \citep{SeymourETAL:16,SingerETAL:16}. 
Such consensus definitions are widely accepted as the basis of ICD codes and build on standard clinical measurements in EHR databases. 
For example, the Sepsis-3 label is assigned by determining an infection accompanied by an organ function deterioration according to the SOFA (Sepsis-related Organ Failure Assessment) criteria \citep{VincentETAL:96}, which are themselves based on thres\-holding various fundamental clinical measurements. The time of sepsis onset is determined by the onset of salient organ dysfunction or the onset of suspicion of infection, or by the earlier of these two events \citep{CohenETAL:24}. Examples for early prediction of sepsis using this automatic labeling scheme are the approaches described in the overview papers of \cite{ReynaETAL:19} or \cite{MoorETAL:21}. 

\begin{figure*}[t!]
    \centering
    \includegraphics[width=0.6\linewidth]{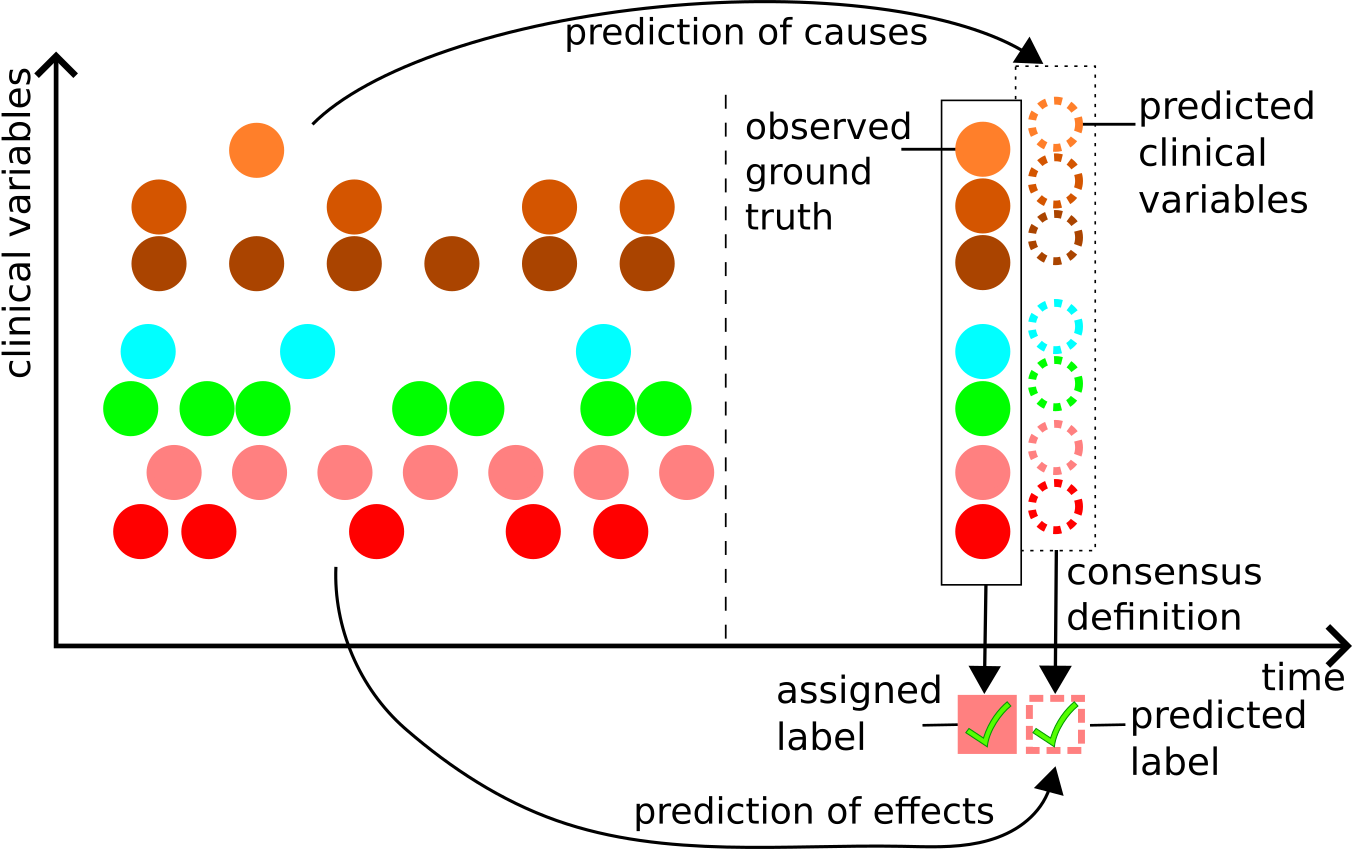}
    \caption{Based on clinical variables observed several hours before the prediction time point (part left of the dashed line), a prediction of effects (bottom arrow) aims to predict the label that the consensus definition assigns to unseen future observations. In contrast, a prediction of causes (top arrow) directly predicts future values of clinical variables and then deterministically applies the consensus definition to these predictions.} 
    \label{fig:prediction}
\end{figure*}

From a machine learning perspective, 
the setup of predicting a label that is the outcome of a consensus definition can be abstractly viewed as the \emph{prediction of an effect}  that is caused by applying a consensus definition to future values of clinical variables.
We propose to turn this setup around:  Instead of predicting an effect based on measurements taken at an earlier time than the values causing the ground truth label, we \emph{directly predict the causes} by forecasting clinical variables, and determine the effect by applying the consensus definition to the forecasted values. This setup is illustrated in Figure \ref{fig:prediction}. 

Prediction of causes has a different goal than prediction of effects in that it aims at the general task of forecasting clinical variables to which arbitrary consensus definitions can be applied, without being restricted to a specific definition that is used to annotate ground truth labels. This flexibility comes at the cost of constituting an arguably harder task since several irregularly and sparsely sampled clinical variables need to be predicted instead of a single non-sparse effect label. However, despite its challenges, it has several intrinsic benefits. First and foremost, a prediction of clinical causes has the invaluable advantage of being immediately interpretable by clinical practitioners. Instead of requiring post-hoc interpretability techniques to understand a SOFA or Sepsis prediction, the predicted clincial causes determining the effect of SOFA or Sepsis can be directly inspected. Furthermore, if a consensus definition is  applied to predicted values of clinical variables, there is no room for shortcut learning \citep{GeirhosETAL:20}, information leakage \citep{KaufmanETAL:11} or circularity \citep{HagmannETAL:23,RiezlerHagmann:24} that can skew a machine learning task. Last, the deterministic application of a consensus definition directly reveals aspects like the choice of onset time that can have a great impact on predictive performance \citep{CohenETAL:24}. It allows a direct investigation of such subtle variations  during testing, while such variations are much harder to explore in approaches where consensus definitions are hard-coded and hidden in the training labels. 

In the following, we present an exemplification of prediction of causes by means of long-term time series forecasting (TSF) of clinical variables, with a focus on accurate prediction of the sparse clinical variables involved in the SOFA-based Sepsis-3 definition \citep{SeymourETAL:16,SingerETAL:16} and the new Simplified Acute Physiology Score (SAPS-II) \citep{LeGallETAL:93} definition.
We present an extensive investigation of optimal network architectures and training procedures to perform long-term clinical TSF on two critical care databases, MIMIC-III \citep{JohnsonETAL:16} and eICU \citep{PollardETAL:18}. We build upon the expressive family of Transformers \citep{VaswaniETAL:17} that has been shown to be able to model dynamical or even chaotic systems \citep{GeshkovskiETAL:23,InoueETAL:22} and has been applied competitively to various TSF tasks \citep{AhmedETAL:23,WenETAL:23}. 
We compare \emph{sparse encoders} specialized to modeling irregularly sampled time series data as sets of observation triplets, containing clinical variable, time of measurement, and measurement value \citep{TipirneniReddy:21} to a standard encoding of time series as a \emph{dense matrix} of input features times time-steps, using binning and mean-value imputation. These encoders are combined with a \emph{direct multi-step (DMS)} forecasting decoder \citep{ZhouETAL:21,WuETAL:21,ZengETAL:23} that has been proposed specifically for long-term TSF, and a standard autoregressive Transformer decoder performing long-term TSF in an \emph{iterative multi-step (IMS)} fashion. Our experiments show that contrary to recent proposals, best results are achieved by a combination of a standard dense encoder with an IMS decoder. This can be attributed to a training strategy called \emph{student forcing} that supplies the model's own previously predicted time steps as context for next time step prediction. Student forcing outperforms the generic \emph{teacher forcing} which relies solely on ground truth contexts \citep{WilliamsZipser:89} as well as scheduled sampling \citep{BengioETAL:15,TeutschMaeder:23} which mixes both context types. 

\begin{figure*}[t!]
    \centering
    \includegraphics[width=0.6\linewidth]{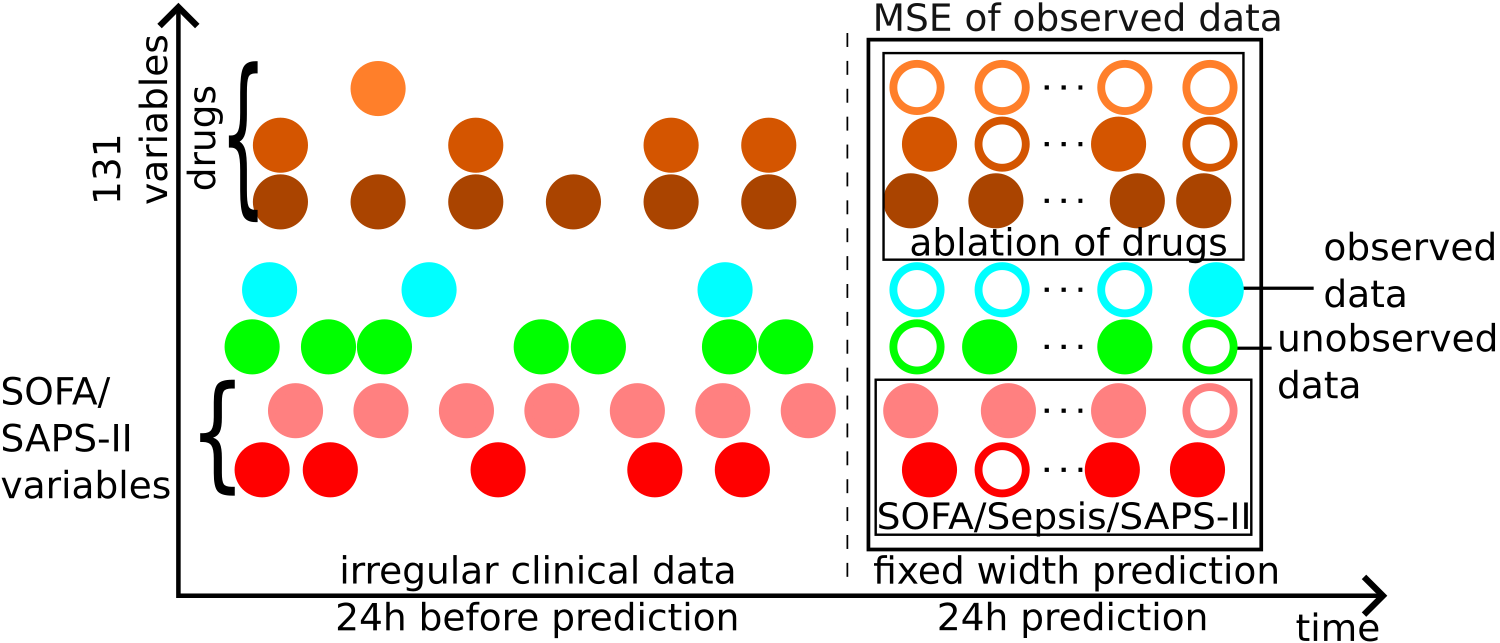}
    \caption{Evaluation setup for TSF experiments: Given irregularly distributed real-world clinical input data, we predict values for 24 hours in the future. We evaluate MSE loss on all variables, SOFA, SAPS-II, and Sepsis-3 accuracy, and perform a study on cross-variate effects of drugs.} 
    \label{fig:io_ou_construction}
\end{figure*}

We present a thorough evaluation of all combinations of the encoders and decoders implemented by us, together with an evaluation of Informer \citep{ZhouETAL:21}, Autoformer \citep{WuETAL:21}, and (D)Linear \citep{ZengETAL:23} models, on progressive and increasingly complex TSF tasks on clinical data. As shown in Figure \ref{fig:io_ou_construction}, our evaluation extends the standard mean squared error (MSE) calculation of 131 clinical variables on the MIMIC-III dataset and 98 clinical variables on the eICU dataset 
to an evaluation of the influence of forecasting accuracy on downstream clinical tasks such as early prediction of SOFA \citep{VincentETAL:96}, Sepsis-3 \citep{SeymourETAL:16,SingerETAL:16}, and SAPS-II \citep{LeGallETAL:93}. Furthermore, we present a study of cross-variate effects of drug administration on other clinical variables.

\subsection*{Generalizable Insights about Machine Learning in the Context of Healthcare}

The contributions of our work to machine learning in the context of healthcare are as follows:

\begin{itemize}
    \item We present a method for early syndrome prediction that is straightforwardly interpretable by medical practitioners. It directly predicts the clinical causes of a diagnosis by long-term TSF of clinical variables, and identifies the label by applying the known consensus definition that has been used to determine the ground truth  to the forecasted clinical measurements. This technique is general and flexible since it is applicable to arbitrary consensus-based prediction tasks.
    \item From a machine learning perspective on long-term TSF, we find that contrary to recent proposals which advocate set function encoders for time series and direct multi-step decoders, best results are achieved by a combination of standard dense encoders with iterative multi-step decoders. The latter can be attributed to a student forcing training strategy that supplies the model’s own predictions as context for next time step prediction, and outperforms standard teacher forcing or scheduled sampling. 
    \item Our experiments are conducted on two critical care databases for SOFA-based Sepsis-3 prediction and prediction of the SAPS-II score. We find consistent wins for the combination of dense encoders and iterative multi-step decoders for both consensus-based prediction tasks on both datasets. 
\end{itemize}

Code and data of this work are available at \url{https://github.com/StatNLP/mlhc_2024_prediction_of_causes}.

\section{Related Work}
\label{sec:related}

TSF of clinical variables is a challenging task in itself. First, clinical data constitute multivariate time series that are irregularly sampled, both in time and across dimensions. Irregular sampling in time series is commonly addressed by data imputation methods. These include simple imputation of unobserved values by last observed or mean values, or by learned values using machine learning, e.g., regression of unobserved values against observed neighboring values (see \cite{FangWang:20} for an overview). Recently, neural ODEs have been proposed as a natural solution for irregular sampling in time series by integrating the dynamics over the time interval \citep{ChenETAL:18,DeBrouwerETAL:19,KidgerETAL:20}. While these techniques show impressive results, they are incompatible with the parallelization idea of Transformers that explicitly avoid recurrence in the time dimension. For our work, we compare the so-called set functions embedding technique \citep{HornETAL:20,TipirneniReddy:21} that encodes time series as sets of observations with a dense encoder that compresses the input time series into 24 hourly bins and uses mean-value imputation. 

Second, clinical patient data require long-term TSF capabilities from machine learning models. Recently, there has been a surge of works on long-term TSF using Transformers  \citep{ZhouETAL:21,WuETAL:21,ZhouETAL:22} and work that explicitly questions the effectiveness of transformers for long-term TSF by conjecturing that the transformer's input permutation invariance may cause an ignorance of temporal input relations \citep{ZengETAL:23}. 
The proposed approaches agree in their argumentation that long-term TSF requires sophisticated DMS techniques in order to overcome the error propagation of IMS techniques. We show that a standard autoregressive Transformer decoder can be tuned to achieve comparable and sometimes even better long-term behavior than DMS decoders. The crucial ingredient is here to teach the Transformer to trust its own predictions as context in prediction of the next time step. We call this training technique student-forcing to differentiate it from teacher-forcing \citep{WilliamsZipser:89}. Combinations of both techniques by scheduled sampling have been shown to be useful in the context of sequence learning and TSF with RNNs \citep{BengioETAL:15,RanzatoETAL:16,TeutschMaeder:23}, however, an investigation of scheduled sampling for TSF with autoregressive Transformers has so far been missing.  

Third, multivariate clinical time series exhibit dependencies such as administration of certain medications resulting in changes in related clinical variables. This requires accurate modeling of cross-variate dependencies between clinical variables. It seems intuitive that models like Transformers that explicitly learn cross-variate connections should be more effective than univariate models like those proposed by \cite{ZengETAL:23} or \cite{NieETAL:23}. As shown in \cite{ChenETAL:23}, the advantages of univariate models come into play only for certain types of benchmarks, but not for datasets that contain complex cross-variate information and auxiliary features. We believe that our investigation of cross-variate effects of drug administration on other clinical variables in critical care data proves the ability of Transformers to model complex cross-variate information in real-world clinical data. 

\begin{figure*}[t!]
    \centering
    \includegraphics[width=0.7\linewidth]{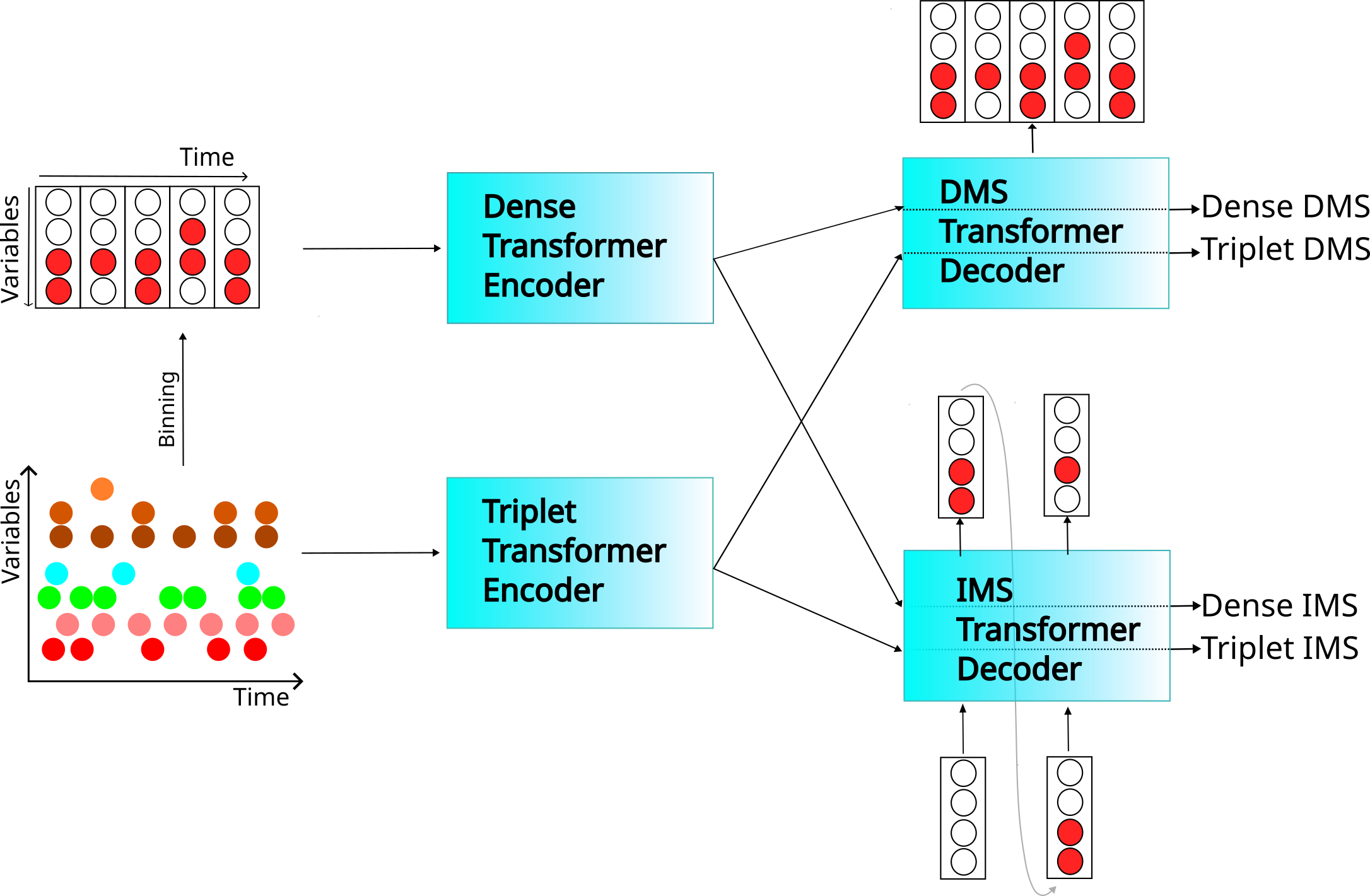}
    \caption{Overview of combination of encoders and decoders at inference time. Irregularly sampled time series (lower left) are encoded by a sparse triplet representation (lower middle), or they are binned into a representation with each vector representing an hour of observations (upper left), and encoded into a dense matrix (upper middle). The encoders are combined with different decoding strategies, either  DMS (non-autoregressive) (upper right) or IMS (autoregressive) (lower right), yielding four different encoder-decoder pipelines.}
    \label{fig:inference_overview}
\end{figure*}

\section{Methods}
\label{sec:methods}

\subsection{Neural Architecture}

Figure \ref{fig:inference_overview} gives an overview of the neural architectures used in our experiments to encode irregularly sampled time series and performing long-term TSF. 

A sparse multivariate input time series comprised of $n$ irregular measurements of $|F|$ clinical variables (lower left) is first variable-wise standardized, and then represented by $n$ triplets $\text{S} = \{(t_i, f_i, v_i)\}_{i=1}^n$, where $t_i \in \mathbb{R}_{\geq 0}$ is a time index, $f_i \in F$ is a clinical variable identifier, and $v_i \in \mathbb{R}$ the observed value of $f_i$ at $t_i$.
For sparse triplet (set function) encoding, each component of a triplet --- time, variable, and value --- receives a separate embedding of length $m$. These embeddings are summed up for each triplet, and an $m \times n$ matrix is fed into a Transformer architecture for encoding (lower middle), following the setup of \cite{TipirneniReddy:21}. 

Alternatively to this sparse triplet encoder, we implement a standard dense encoding architecture.
To achieve this, the irregularly sampled input data (lower left) is binned into hourly buckets by recording the first observed value of each variable (upper left) and applying mean imputation for unobserved values (effectively resulting in imputation of zeros because of standardization). Furthermore, we generate a masking matrix of the same size that informs the model if a corresponding value in the binned matrix is imputed or observed and append this matrix to the first one. 
A time series of length $D := \max(t_i)$\footnote{For simplicity, we assume that $t_i$ is measured in hours.} then results in a dense $\lceil D \rceil \times 2|F|$ matrix that is fed into a standard Transformer encoder \citep{VaswaniETAL:17} (upper middle). 

We implement two decoder types to generate dense target time series of length $T$ for an encoded input $x$. The direct multi-step (DMS) decoder uses $T$ randomly initialized self-attention modules to perform $T$ prediction steps at once. This non-autoregressive decoder predicts time steps independent of each other, and the inferred output $\hat{y}_t$  depends only on the encoded input $x$ and the model parameters $\theta$:
\begin{align}
\label{eq:no autoreg}
    \hat{y}_t = f_\theta(x)
\end{align}
This method is depicted as the DMS strategy (upper right) in Figure \ref{fig:inference_overview}. 

Alternatively to the DMS decoder, the iterative multi-step (IMS) decoder generates an output vector $\hat{y}_t \in \mathbb{R}^{|F|}$ using a standard autoregressive model \citep{VaswaniETAL:17}. The inferred output $\hat{y}_t$ is a function of the history $\hat{y}_{<t}$ of predicted tokens until time $t$, the encoded input $x$, and the model parameters $\theta$:
\begin{align}
\label{eq:autoreg}
    \hat{y}_t = f_\theta(\hat{y}_{<t}, x)
\end{align}
To perform long-term TSF using the IMS setup, the outputs $\hat{y}_t$ from each time step $t = 1, \ldots, T$ are concatenated, yielding the IMS strategy (lower right) in Figure \ref{fig:inference_overview}.

\begin{figure}[t!]
    \centering
    \includegraphics[width=0.7\linewidth]{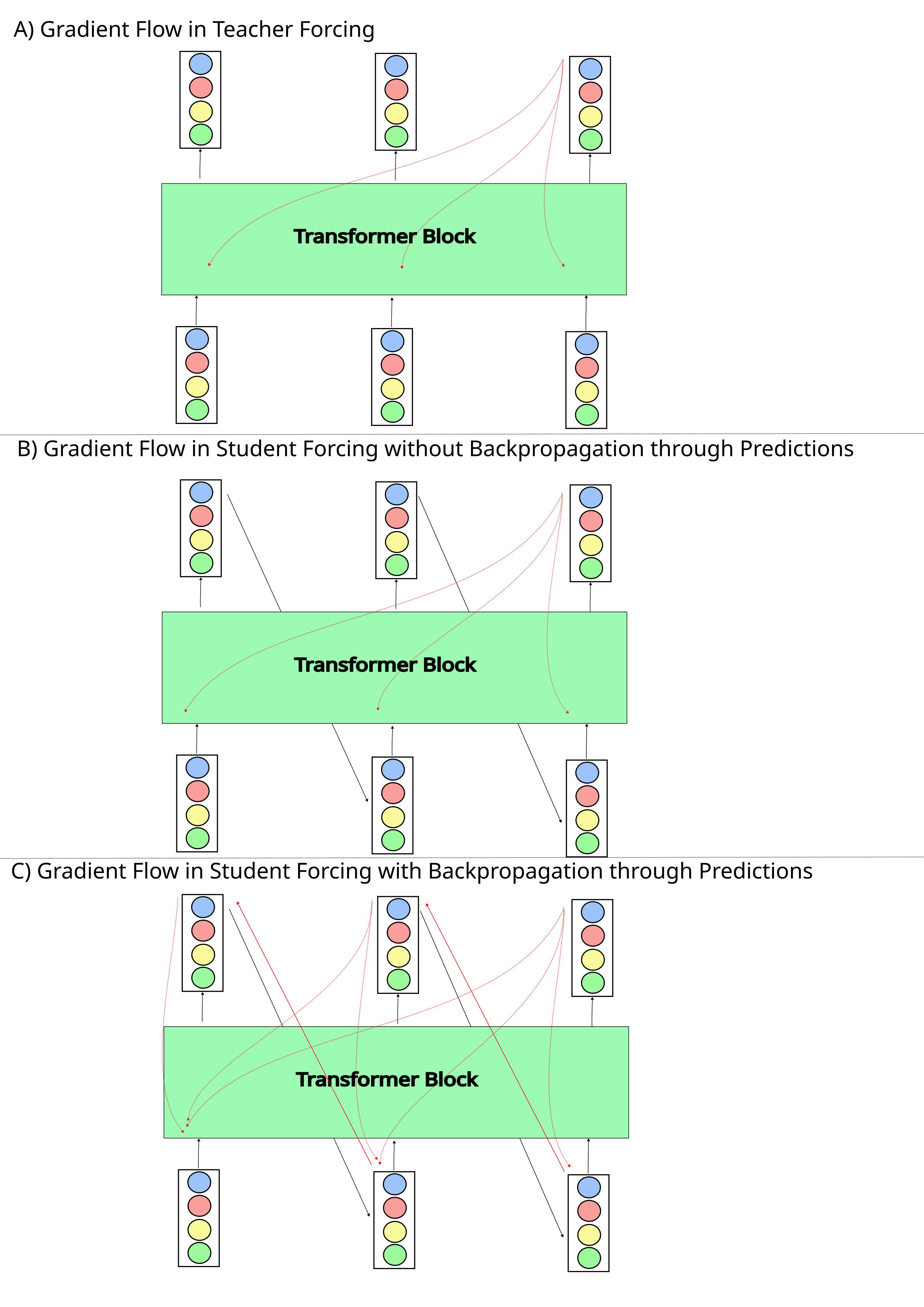}
    \caption{Gradient flow in time for different styles of training. Gradients (in red) are only shown for a loss generated at the third timestep.}
    \label{fig:gradient_flow}
\end{figure}

\subsection{Training Methods}

It is hypothesized in the literature that DMS decoders are superior to IMS decoders for long-term TSF because of potential error propagation in the latter \citep{ZhouETAL:21,WuETAL:21,ZengETAL:23}. In our experiments, we show that an IMS decoder can be tuned to achieve comparable and even better long-term TSF behavior than DMS decoders by applying a strategy called \emph{student-forcing} during training. Student forced training of an autoregressive decoder means to apply equation \ref{eq:autoreg} during training in contrast to \emph{teacher-forcing} \citep{WilliamsZipser:89} where during training the history for next-token prediction is the ground truth value $y_{<t}$:
\begin{align}
    \hat{y}_t = f_\theta(y_{<t}, x)
\end{align}

Additionally, we combine teacher forcing and student forcing via scheduled sampling \citep{BengioETAL:15,TeutschMaeder:23}. Following \cite{TeutschMaeder:23}, we implemented linear curricula determining the teacher forcing ratio $C_{lin}(e)$ per epoch $e$ as 
\begin{equation}
    C_{lin}(e) = \epsilon_{start}+(\epsilon_{end}-\epsilon_{start})\cdot\frac{\min(e, \text{\L})}{\text{\L}}.
\end{equation}
We varied between an increasing ($\epsilon_{start}=0.25$ and $\epsilon_{end}=1$) and a decreasing curriculum ($\epsilon_{start}=1$ and $\epsilon_{end}=.25$), both with a curriculum length \text{\L} of 200. We also varied the selection of time steps for teacher forcing, either in a randomized or deterministic way. When teacher forcing is performed randomly, a time step is teacher forced with probability $C_{lin}(e)$. When time steps are selected in a deterministic way, teacher forcing depends not only on $C_{lin}(e)$, but also on the current position within the predicted sequence of length $l$ where only the first $\lfloor l \cdot C_{lin}(e)\rfloor$ positions are teacher forced.

Different to natural language processing, the Transformer architecture, when applied to TSF, is used to predict continuous values. This allows us to enhance the expressive power of the model by back-propagating through the continuously differentiable prediction history. The gradient flow of teacher forcing, student forcing, and student forcing with backpropagation through predictions is shown in Figure \ref{fig:gradient_flow}. As shown in our experiments, this modified gradient flow improves the model further.

\begin{table*}[t!]
	\centering
	\small
	\caption{Main evaluation results on MIMIC-III for combinations of sparse and dense encoders with direct multi-step (DMS) and iterative multi-step (IMS) decoders, trained by teacher forcing (TF), student forcing (SF), or scheduled sampling (SS), with or without backpropagation (BP) through predictions. Curricula for scheduled sampling are denoted as deterministic increasing (DI) or decreasing (DD), randomized increasing (RI) or decreasing (RD). Evaluation is done according to MSE of all variables (Equation \ref{eq:mse}), MSE for SOFA (Equation \ref{eq:SOFA-MSE}), and Accuracy for Sepsis (Equation \ref{eq:sepsis_acc}). Numbers in subscripts denote the 95\% confidence interval for the estimation of the respective evaluation score on the test set. Best results are shown in bold face.}
	\label{tab:main_results}
	\begin{tabular}{lllllccc}
		\toprule
		Model      & Enc & Dec & Train& BP  &            MSE             &          MSE-SOFA          &                 Acc-Sepsis                  \\ \midrule
1&Triplet&DMS&-&-&7.326$_{[7.320,7.332]}$&2.825$_{[2.822,2.828]}$&90.11$_{[88.08,92.14]}$\\
2&Triplet&IMS&TF&-&12.371$_{[12.360,12.382]}$&10.785$_{[10.776,10.794]}$&75.14$_{[72.19,78.08]}$\\
3&Triplet&IMS&SF&No&6.881$_{[6.875,6.887]}$&2.872$_{[2.869,2.876]}$&89.56$_{[87.48,91.64]}$\\
4&Triplet&IMS&SF&Yes&7.065$_{[7.058,7.073]}$&2.917$_{[2.912,2.922]}$&89.29$_{[87.18,91.39]}$\\
5&Triplet&IMS&SS-DI&No&8.764$_{[8.753,8.775]}$&3.466$_{[3.457,3.475]}$&82.85$_{[80.28,85.42]}$\\
6&Triplet&IMS&SS-DI&Yes&8.993$_{[8.982,9.004]}$&4.150$_{[4.141,4.159]}$&80.43$_{[77.73,83.14]}$\\
7&Triplet&IMS&SS-DD&No&24.360$_{[24.349,24.371]}$&16.345$_{[16.336,16.353]}$&70.65$_{[67.55,73.75]}$\\
8&Triplet&IMS&SS-DD&Yes&22.046$_{[22.035,22.057]}$&11.055$_{[11.046,11.063]}$&63.53$_{[60.25,66.81]}$\\
9&Triplet&IMS &SS-RI&No&7.068$_{[7.062,7.073]}$&2.689$_{[2.685,2.693]}$&85.63$_{[83.24,88.02]}$\\
10&Triplet&IMS&SS-RI&Yes&6.967$_{[6.960,6.973]}$&2.609$_{[2.605,2.613]}$&86.47$_{[84.14,88.80]}$\\
11&Triplet&IMS &SS-RD&No&7.761$_{[7.752,7.771]}$&3.092$_{[3.085,3.100]}$&86.11$_{[83.76,88.47]}$\\
12&Triplet&IMS& SS-RD&Yes&7.389$_{[7.382,7.395]}$&2.857$_{[2.852,2.862]}$&86.71$_{[84.40,89.03]}$\\
\midrule
13&Dense&DMS&-&-&8.789$_{[8.780,8.798]}$&6.024$_{[6.016,6.031]}$&85.87$_{[83.50,88.24]}$\\
14&Dense&IMS&TF&-&11.907$_{[11.896,11.918]}$&9.242$_{[9.233,9.251]}$&66.18$_{[62.96,69.41]}$\\
15&Dense&IMS&SF&No&7.456$_{[7.447,7.466]}$&3.376$_{[3.368,3.384]}$&88.96$_{[86.66,91.25]}$\\
16&Dense&IMS&SF&Yes&\textbf{6.216}$_{[6.211,6.221]}$&\textbf{2.497}$_{[2.495,2.500]}$&\textbf{90.34}$_{[88.33,92.35]}$\\
17&Dense&IMS&SS-DI&No&8.861$_{[8.850,8.871]}$&3.598$_{[3.590,3.607]}$&83.09$_{[80.54,85.64]}$\\
18&Dense&IMS&SS-DI&Yes&9.159$_{[9.148,9.170]}$&4.008$_{[3.999,4.017]}$&80.19$_{[77.48,82.91]}$\\
19&Dense&IMS&SS-DD&No&10.466$_{[10.455,10.478]}$&6.392$_{[6.382,6.401]}$&75.60$_{[72.68,78.53]}$ \\
20&Dense&IMS&SS-DD&Yes&10.513$_{[10.502,10.524]}$&6.422$_{[6.413,6.431]}$&76.57$_{[73.68,79.46]}$\\
21&Dense&IMS&SS-RI&No&6.929$_{[6.924,6.935]}$&2.830$_{[2.826,2.834]}$&86.47$_{[84.14,88.80]}$\\
22&Dense&IMS&SS-RI&Yes&6.909$_{[6.904,6.915]}$&2.770$_{[2.766,2.773]}$&85.27$_{[82.85,87.68]}$\\
23&Dense&IMS&SS-RD&No&11.749$_{[11.738,11.760]}$&9.667$_{[9.658,9.676]}$&64.86$_{[61.60,68.11]}$\\
24&Dense&IMS&SS-RD&Yes&11.697$_{[11.686,11.709]}$&9.525$_{[9.516,9.534]}$&64.37$_{[61.11,67.63]}$\\
\midrule
Informer&Dense&DMS&-&-&6.354$_{[6.350,6.358]}$&2.616$_{[2.614,2.617]}$&85.75$_{[83.37,88.13]}$\\
Autoformer&Dense&DMS&-&-&7.438$_{[7.432,7.444]}$&3.352$_{[3.348,3.356]}$&89.61$_{[87.54,91.69]}$\\
DLinear&Dense&DMS&-&-&8.487$_{[8.479,8.495]}$&5.019$_{[5.012,5.025]}$&76.81$_{[73.94,79.69]}$\\
Linear&Dense&DMS&-&-&8.486$_{[8.477,8.494]}$&5.001$_{[4.995,5.007]}$&77.29$_{[74.44,80.15]}$\\
\bottomrule
	\end{tabular}
\end{table*}

\begin{table*}[t!]
	\centering
	\small
	\caption{Evaluation results on MIMIC-III showing MSE at beginning 8 hours (MSE$_{1:8}$) and last 16 hours (MSE$_{9:24}$) of 24 hour forecasting period. Numbers in subscript in square brackets denote the 95\% confidence interval for the estimation of the respective evaluation score on the test set. Best results are shown in bold face.}
	\label{tab:partial_mse_results}
	\begin{tabular}{lllllcc}
		\toprule
		Model      & Enc & Dec & Train & BP  &              MSE$_{1:8}$              &             MSE$_{9:24}$              \\ \toprule
1&Triplet&DMS&-&-&7.687$_{[7.676,7.697]}$&7.160$_{[7.154,7.165]}$\\
2&Triplet&IMS&TF&-&12.753$_{[12.734,12.773]}$&12.203$_{[12.195,12.212]}$\\
3&Triplet&IMS&SF&No&6.686$_{[6.676,6.696]}$&6.991$_{[6.986,6.997]}$\\
4&Triplet&IMS&SF&Yes&6.899$_{[6.886,6.911]}$&7.162$_{[7.156,7.168]}$\\
5&Triplet&IMS&SS-DI&No&9.335$_{[9.316,9.354]}$&8.495$_{[8.487,8.503]}$\\
6&Triplet&IMS&SS-DI&Yes&9.593$_{[9.574,9.612]}$&8.710$_{[8.702,8.718]}$\\
7&Triplet&IMS &SS-DD&No&28.394$_{[28.374,28.413]}$&22.390$_{[22.382,22.399]}$\\
8&Triplet&IMS &SS-DD&Yes&22.163$_{[22.144,22.183]}$&22.033$_{[22.024,22.041]}$\\
9&Triplet&IMS &SS-RI&No&6.914$_{[6.903,6.925]}$&7.158$_{[7.153,7.162]}$\\
10&Triplet&IMS &SS-RI&Yes&6.705$_{[6.694,6.717]}$&7.110$_{[7.104,7.116]}$\\ 
11&Triplet&IMS &SS-RD&No&7.654$_{[7.638,7.671]}$&7.829$_{[7.822,7.837]}$\\
12&Triplet&IMS &SS-RD&Yes&7.205$_{[7.192,7.217]}$&7.494$_{[7.488,7.500]}$\\
\midrule
13&Dense&DMS&-&-&9.116$_{[9.099,9.133]}$&8.641$_{[8.634,8.648]}$\\
14&Dense&IMS&TF&-&12.627$_{[12.608,12.646]}$&11.570$_{[11.562,11.578]}$\\
15&Dense&IMS&SF&No&7.427$_{[7.410,7.445]}$&7.484$_{[7.477,7.491]}$\\
16&Dense&IMS&SF&Yes&\textbf{6.117}$_{[6.109,6.124]}$&\textbf{6.278}$_{[6.273,6.283]}$\\
17&Dense&IMS&SS-DI&No&9.328$_{[9.309,9.347]}$&8.643$_{[8.635,8.651]}$\\
18&Dense&IMS&SS-DI&Yes&9.744$_{[9.725,9.763]}$&8.884$_{[8.876,8.892]}$\\
19&Dense&IMS&SS-DD&No&11.258$_{[11.239,11.278]}$&10.090$_{[10.082,10.098]}$\\
20&Dense&IMS&SS-DD&Yes&11.331$_{[11.311,11.350]}$&10.124$_{[10.116,10.133]}$\\
21&Dense&IMS&SS-RI&No&6.752$_{[6.742,6.763]}$&7.031$_{[7.027,7.036]}$\\
22&Dense&IMS&SS-RI&Yes&6.587$_{[6.578,6.597]}$&7.084$_{[7.078,7.089]}$\\
23&Dense&IMS&SS-RD&No&12.554$_{[12.535,12.574]}$&11.369$_{[11.361,11.377]}$\\
24&Dense&IMS&SS-RD&Yes&12.575$_{[12.556,12.595]}$&11.281$_{[11.272,11.289]}$\\
\midrule
Informer&Dense&DMS&-&-&6.349$_{[6.343,6.354]}$&6.369$_{[6.365,6.373]}$\\
Autoformer&Dense&DMS&-&-&7.232$_{[7.224,7.241]}$&7.556$_{[7.550,7.561]}$\\
DLinear&Dense&DMS&-&-&8.360$_{[8.346,8.373]}$&8.568$_{[8.561,8.575]}$\\
Linear&Dense&DMS&-&-&8.359$_{[8.345,8.372]}$&8.566$_{[8.559,8.573]}$\\
\bottomrule
	\end{tabular}
\end{table*}

\begin{table*}[t!]
	\centering
	\small
	\caption{SAPS-II evaluation results on MIMIC-III.
 Numbers in subscripts denote the 95\% confidence interval for the estimation of the respective evaluation score on the test set. Best results are shown in bold face.}
	\label{tab:saps_results}
	\begin{tabular}{lllllcc}
		\toprule
		Model      & Enc & Dec & Train& BP  &            MSE             &          MSE-SAPS-II                     \\ \midrule
1&Triplet&DMS&-&-&7.326$_{[7.320,7.332]}$&222.426$_{[222.417,222.435]}$\\
2&Triplet&IMS&TF&-&12.371$_{[12.360,12.382]}$&270.245$_{[270.236,270.254]}$\\
3&Triplet&IMS&SF&No&6.881$_{[6.875,6.887]}$&187.826$_{[187.817,187.835]}$\\
4&Triplet&IMS&SF&Yes&7.065$_{[7.058,7.073]}$&187.826$_{[187.817,187.835]}$\\
5&Triplet&IMS&SS-DI&No&8.764$_{[8.753,8.775]}$&167.347$_{[167.338,167.356]}$\\
6&Triplet&IMS&SS-DI&Yes&8.993$_{[8.982,9.004]}$&166.765$_{[166.756,166.774]}$\\
7&Triplet&IMS&SS-DD&No&24.360$_{[24.349,24.371]}$&147.559$_{[147.550,147.567]}$\\
8&Triplet&IMS&SS-DD&Yes&22.046$_{[22.035,22.057]}$&377.004$_{[376.996,377.013]}$\\
9&Triplet&IMS&SS-RI&No&7.068$_{[7.062,7.073]}$&102.445$_{[102.442,102.449]}$\\
10&Triplet&IMS&SS-RI&Yes&6.967$_{[6.960,6.973]}$&110.174$_{[110.170,110.178]}$\\
11&Triplet&IMS&SS-RD&No&7.761$_{[7.752,7.771]}$&101.756$_{[101.749,101.764]}$\\
12&Triplet&IMS&SS-RD&Yes&7.389$_{[7.382,7.395]}$&103.756$_{[103.751,103.761]}$\\
\midrule
13&Dense&DMS&-&-&8.789$_{[8.780,8.798]}$&132.819$_{[132.811,132.826]}$\\
14&Dense&IMS&TF&-&11.907$_{[11.896,11.918]}$&183.693$_{[183.684,183.702]}$\\
15&Dense&IMS&SF&No&7.456$_{[7.447,7.466]}$&109.381$_{[109.373,109.389]}$\\
16&Dense&IMS&SF&Yes&\textbf{6.216}$_{[6.211,6.221]}$&\textbf{89.115}$_{[89.112,89.117]}$\\
17&Dense&IMS&SS-DI&No&8.861$_{[8.850,8.871]}$&169.390$_{[169.381,169.399]}$\\
18&Dense&IMS&SS-DI&Yes&9.159$_{[9.148,9.170]}$&175.625$_{[175.616,175.634]}$\\
19&Dense&IMS&SS-DD&No&10.466$_{[10.455,10.477]}$&192.777$_{[192.768,192.786]}$\\
20&Dense&IMS&SS-DD&Yes&10.513$_{[10.502,10.524]}$&166.208$_{[166.199,166.217]}$\\
21&Dense&IMS&SS-RI&No&6.929$_{[6.923,6.934]}$&107.428$_{[107.424,107.432]}$\\
22&Dense&IMS&SS-RI&Yes&6.909$_{[6.904,6.915]}$&105.098$_{[105.095,105.101]}$\\
23&Dense&IMS&SS-RD&No&10.466$_{[10.455,10.478]}$&169.432$_{[169.423,169.440]}$\\
24&Dense&IMS&SS-RD&Yes&10.513$_{[10.502,10.524]}$&206.098$_{[206.089,206.107]}$\\
\midrule
Informer&Dense&DMS&-&-&6.354$_{[6.350,6.358]}$&92.008$_{[92.006,92.010]}$\\
Autoformer&Dense&DMS&-&-&7.438$_{[7.432,7.444]}$&95.306$_{[95.302,95.310]}$\\
DLinear&Dense&DMS&-&-&8.487$_{[8.479,8.495]}$&129.476$_{[129.470,129.482]}$\\
Linear&Dense&DMS&-&-&8.486$_{[8.477,8.494]}$&129.570$_{[129.563,129.576]}$\\
              \bottomrule
	\end{tabular}
\end{table*}      

\section{Experiments}
\label{sec:exps}

\subsection{Datasets}

In our experiments, we use the Medical Information Mart for Intensive Care III (MIMIC-III) data \citep{JohnsonETAL:16}. The data were collected from the Beth Israel Deaconess Medical Center between 2001 and 2012 and contain over 
40k patients. After filtering for patients with an ICU stay of at least 24 hours with reported gender and age of at least 18 years, our dataset contains 44,858 ICU stays with 56 million data points. We split the data into partitions for training (28,708), development (7,270), and testing (8,880). For our study, we used 131 different clinical variables, including 15 clinical markers for the calculation of SOFA and 13 for SAPS-II, and the demographic variables gender and age. This selection comprises all vital signs and laboratory values used in the PhysioNet challenge for early prediction of sepsis \citep{ReynaETAL:19}. The full list of extracted MIMIC-III features is given in Appendix \ref{appendix:features}. As shown in Appendix \ref{appendix:histograms}, the sparsity of the dataset is very high, meaning that the average number of observations per patient per hour is at most 20 and rapidly declines with longer length of stay. 
Converting the data to a dense one hour representation yields 89.08\% missing data, changing per variable from under 15\% (HR, RR, SBP, DBP, MBP, and O2 Saturation) to more than 90\% for 101 variables, and exceeding 99\% for 42 variables. On the other side we are loosing 17.73\% of the data points through the densification procedure. Demoscopic data are complete for all patients.

Furthermore, we conducted some experiments on the larger eICU dataset \citep{PollardETAL:18}. These data were collected from over 200 US hospitals and comprise over 200,000 ICU stays. After filtering for patients with an ICU stay of at least 48 hours, reported gender and aged 18 years or older, 
we arrived at 77,704 ICU stays with 415 million datapoints. This set was partitioned in subsets for training (49,730), development (12,433), and testing (15,541).  As shown in Appendix \ref{appendix:features}, we extracted 98 clinical variables and 17 demographic markers for our experiments. As shown in Appendix \ref{appendix:histograms}, the measurements in the eICU are denser than in MIMIC-III  since the number of observations per patient per hour is three times higher than for MIMIC-III and decreases at a slower rate with length of stay.
In a dense encoding, this reduces to one sixth of the measurements (16.86\%). To densify the data like before yields even 89.85\% missing data. The same six variables as before are quite complete while 84 variables are missing more than 90\% and 29 variables 99\%. The demographic data are almost complete.

\subsection{Evaluation Measures}

We use the following measures to evaluate our models for long-term TSF. Given $N$ time series in our dataset, with $T$ hours prediction for TSF, the mean squared error (MSE) over hourly prediction vectors $\hat{y}^n_t$ is defined as follows:
\begin{equation}
\label{eq:mse}
\text{MSE}=\frac{1}{NT}\sum_{n=1}^{N}\sum_{t=1}^{T}||(y^n_t-\hat{y}^n_t) \odot m^n_t||^2_2 
\end{equation}
where $m^n_t \in \{0,1\}^{|F|}$  is a mask indicating if the variables in $y^n_t$ were observed or not, and $\odot$ is a component-wise product. In our experiments, $T$ is set to 24 hours.

For a closer inspection of the long-term TSF error, we compute MSE at the beginning and at the end of the forecasting task. We calculate $\text{MSE}_{1:8}$ for the first 8 hours, and $\text{MSE}_{9:24}$ for the last 16 hours of a 24 hour forecasting period:
\begin{align}
\text{MSE}_{1:8}&=\frac{1}{N\cdot8}\sum_{n=1}^{N}\sum_{t=1}^{8}||(y^n_t-\hat{y}^n_t) \odot m^n_t||^2_2, \\
\text{MSE}_{9:24}&=\frac{1}{N\cdot16}\sum_{n=1}^{N}\sum_{t=9}^{24}||(y^n_t-\hat{y}^n_t) \odot m^n_t||^2_2.
\end{align}

Prediction of SOFA is evaluated by the MSE between the SOFA score computed on the ground truth values and on the forecasted variable values relevant for the consensus definition \citep{VincentETAL:96}. The SOFA score is defined as the sum of  six organ system subscores ranging from 0-4 depending for their part on thresholded fundamental clinical variables observed during a 24h window (see Appendix \ref{appendix:sofa}). 
We compared the sub-scores obtained from forecasted data ($\widehat{\text{SOFA}_6}$) to those obtained from the corresponding gold data ($\text{SOFA}_6$) masking forecasted values that have no corresponding observed gold value by the following MSE calculation:
\begin{equation}
\label{eq:SOFA-MSE}
\text{MSE-SOFA}=\frac{1}{N}\sum_{n=1}^{N}||\text{SOFA}_6^n-\widehat{\text{SOFA}_6}^n||^2_2.
\end{equation}

Accuracy of sepsis prediction is evaluated based on the Sepsis-3 consensus definition \citep{SingerETAL:16, SeymourETAL:16}. According to this definition, a patient is classified as septic if two criteria are met. Firstly, a patient must have a verified or suspected infection, and secondly, a SOFA score showing an increase of 2 or more points in the following 24 hours after the infection. Similar to \cite{SeymourETAL:16}, we identify a suspected infection by a combination of antibiotics treatment and blood culture, starting within the first 24 hours after admission, and gather all infected patients in the set $I$. For those patients, $\text{SOFA}_{1:24}$ denotes the SOFA score for the first 24 hours. For the following 24 hours, $\text{SOFA}_{25:48}$ denotes the SOFA score based on the ground truth data, and $\widehat{\text{SOFA}}_{25:48}$ denotes the SOFA score based on the forecasted data. The accuracy of the predicted sepsis label is calculated as the average match with the ground truth label, where labels are assigned by a check whether a change in SOFA $\geq 2$ happens between the first 24 hours and the prediction window of 24 hours. We use the following notation for indicator functions: $[\![ a ]\!] = 1$ if $a$ is true, $0$ otherwise. Accuracy of Sepsis-labeling can then be defined as follows: 
\begin{align}
    \label{eq:sepsis_acc}
        \text{Acc-Sepsis} &= \frac{1}{|I|}\sum_{i \in I} [\![ \chi_i = \hat{\chi}_i ]\!], \\ \notag
    \text{ where } & \chi_i := [\![ (\text{SOFA}^i_{25:48} - \text{SOFA}^i_{1:24}) \geq 2 ]\!],\\ \notag
\text{ and } & \hat{\chi}_i := [\![ (\widehat{\text{SOFA}^i}_{25:48} - \text{SOFA}^i_{1:24}) \geq 2 ]\!]. \notag
\end{align}

The new Simplified Acute Physiology Score (SAPS-II) \citep{LeGallETAL:93} scores the illness severity of an ICU patient based on a moderate number of routine clinical measurements collected during a 24 hour period. Our implementation is reduced to all non-static variables involved in the calculation of SAPS-II scores (see Appendix \ref{appendix:sapsii}).
To evaluate predictions for SAPS-II, we calculate the MSE between the integer point scores ranging between 0 and 120 for ground truth SAPS-II and predicted $\widehat{\text{SAPS-II}}$ scores:\footnote{We ignore the standard probability conversion for the MSE calculation since the comparison of untransformed scores provides a more accurate evaluation of the utility of the forecasted data.} 
\begin{equation}
\label{eq:SAPS-MSE}
\text{MSE-SAPS-II}=\frac{1}{N}\sum_{n=1}^{N}(\text{SAPS-II}^n-\widehat{\text{SAPS-II}}^n)^2
\end{equation}

\subsection{Experimental Results}

All training runs in our experiments used a 24 hour observation window, followed by a 24 hour prediction window. In order to best exploit the training data, we used sliding windows of 24 hours observations and 24 hours prediction that were shifted in 4 hour steps from admission time up to five days. The learning objective is to minimize the MSE by applying Equation \ref{eq:mse} to the training data. Extensive metaparameter search was conducted on the MIMIC-III and eICU development sets (see Appendix \ref{appendix:sofa}) where the best result on the development set was chosen for final evaluation on the test set.

Table \ref{tab:main_results} shows the main results of our evaluation of various combinations of encoders and decoders, and competing current approaches, on increasingly complex TSF tasks. The first task is the standard evaluation of the MSE (Equation \ref{eq:mse}) for TSF of 131 clinical variables on the MIMIC-III testset. 
 The best result is obtained for model 16 --- a dense encoder, combined with an IMS decoder, trained with student forcing and backpropagation. 
In general, student forcing is far better than teacher forcing (models 2 and 14) or scheduled sampling, which underperforms especially when a decreasing schedule is used (models 7, 8, 19, 20, 23, 24), irrespective of the encoding method. 
Our evaluation progresses from MSE of all clinical variables to a subset of 15 clinical variables that are relevant for the computation of the SOFA score (see the variables marked with $^*$ in Appendix \ref{appendix:features}). MSE-SOFA is computed according to Equation \ref{eq:SOFA-MSE}. Table \ref{tab:main_results} shows that MSE-SOFA is well correlated with MSE over all variables, with similar rankings in both columns. Best results are again obtained for the combination of a dense encoder with an IMS decoder, trained with student forcing and backpropagation (model 16). 
The last column in Table \ref{tab:main_results} presents the accuracy of Sepsis prediction. The best results are obtained again for model 16. Since Sepsis accuracy is computed according to Equation \ref{eq:sepsis_acc}, by filtering patients with infection, the data filtering resulted in a small test set of only 345 patients. Because of the high variance in evaluation scores, many confidence intervals overlap, so that we cannot conclude statistical significance of result differences. In Table \ref{tab:eicu_main_results} in Appendix \ref{appendix:eicu}, we report similar results on the larger eICU dataset. There the number of patients with suspected infection is larger (3,789 patients), and the confidence intervals are accordingly smaller. Furthermore, in Appendix \ref{appendix:f1}, we report a breakdown of Sepsis prediction results according to true positives, false positives, true negative, false negatives, and F1 score. Model 16 outperforms all competitors in these tables as well.
Out of the competing models shown in the last block of Table \ref{tab:main_results}, the Informer model achieves a competitive MSE and MSE-SOFA, however, these advantages do not carry over to Sepsis accuracy comparable with model 16. 

The results presented in Table \ref{tab:partial_mse_results} allow us to draw a more nuanced comparison between student- and teacher-forcing. We see indications of error propagation for student-forcing, shown by a slight MSE increase for later time-steps for models 3, 4, 15, and 16. However, this effect is magnitudes smaller than the beneficial consequences of student-forcing. Teacher-forced models 2 and 14 does not suffer from error propagation. The interpolation between pure student- and teacher-forcing that is done via scheduled sampling shows that teacher forcing is exceedingly detrimental in the early stages of training and when applied non-randomly. Thus, scheduled sampling cannot offer a way to mitigate the small error propagation happening in student forcing.

An evaluation of the prediction of the SAPS-II prediction is given in Table \ref{tab:saps_results}. The ranking of models according to MSE results for all variables is similar to the ranking in Table \ref{tab:main_results}, and consistent with the prediction of SOFA and Sepsis, the combination of a dense encoder with an iterative multi-step decoder and backpropagation through predictions achieves the best results. 

Similar experiments were conducted on the eICU dataset (full result tables are given in Appendix \ref{appendix:eicu}). Consistent with the results on MIMIC-III, best results for MSE-SOFA and accuracy of Sepsis prediction are obtained for the combination of a dense encoder with an IMS decoder, trained with student forcing and backpropagation (model 16). This model also achieves best results for SAPS-II prediction. Since eICU  is larger than MIMIC-III, confidence intervals are smaller, yielding increased significance of result differences.

Lastly, we performed an experiment where compared direct prediction of the SOFA score (called effect prediction) our best TSF model (called prediction of causes). We conducted this experiment by adding a regression head to our dense encoder, and trained this model on automatically assigned SOFA scores (ranging from 0 to 24) in order to use the whole training set. The evaluation was done by computing MSE of the predicted scores.
The resulting MSE scores on eICU were 2.2869 for model 16 compared to 2.6748 for the regression model. On MIMIC-III, we obtained MSE scores of 2.6973 for model 16 and of 2.5865 for the regression model. This shows that prediction of causes is comparable to prediction of effects. A possible explanation for the slightly worse result of direct effect prediction on eICU is the high sparsity of target variables for prediction of causes, leading to many masked values that are ignored in the evaluation of the latter approach. Masking for TSF is consistent with the SOFA consensus definition where unobserved values are set to default values of 0, and with clinical practice where data are collected in a panel-wise fashion with dedicated time slots causing sparsity of observed data.

\subsection{Cross-variate Effects of Cardio-vascular Medications}

In order to assess the cross-variate effects of cardio-vascular drug administration for our best model on MIMIC-III (model 16), we created synthetic inputs based on test data where we separately altered the value of dopamine, dobutamine, and norepinephrine during decoding, keeping everything else unchanged. For each drug, we decoded an input two times, setting the respective drug level to the first (Q1) and third quartile (Q3) of sampled drug doses during decoding for each time step. For analysis, the resulting predictions were averaged over 24 time points for all variables (except for the manipulated drug, which was excluded from analysis), and the two  groups (Q1 and Q3) were compared by t-tests  and Mann-Whitney-U-Test applying the Bonferroni correction \citep{Bonferroni:35} to account for multiple testing. There are only two statistically significant results in both tests and their effect size according to Cohen's $d$ \citep{Cohen:13} is reported.
The results of this ablation of drugs show that model 16 has learned that larger doses of dobutamine are associated with lower doses of amidarone  ($d=.11$). This result is in line with a contraindication of dobutamine for patients that suffer from arrhythmia. We also observe a small positive association ($d=.09$) between dobutamine dose and midazolam dose, gastric meds and fiber.

\section{Discussion}

Most machine learning approaches to early syndrome diagnosis define ground truth labels as the effect of an application of a known medical consensus definition to future clinical measurements. Knowledge of this construction of the ground truth suggests to predict the clinical causes to which the known consensus definition can be deterministically applied. This leads to a prediction that is straightforwardly interpretable by clinicians and can be used for arbitrary consensus-based prediction tasks. The machine learning focus is then shifted to accurate long-term TSF of clinical variables that are fundamental to consensus definitions of syndromes. Since consensus definitions such as the SOFA-based Sepsis-3 and SAPS-II scores are mostly based on sparsely observed laboratory measurements, a proper encoding of sparse inputs together with a decoding strategy that exploits dependencies between multivariate inputs is key. 
Our experiments on two datasets show that a combination of a standard dense encoder using data imputation with an iterative multi-step forecasting outperforms specialized set function encoders and direct multi-step decoders. We conjecture that the accuracy advantage of the dense encoder is attributable to the compression of long input time series into 24 hourly bins that record the most important observations, while the advantage of the IMS decoder lies in its ability to capture cross-variate dependencies.

\paragraph{Limitations}

The experiments in our work were conducted by choosing the best metaparameter setting on the development set for final evaluation on the test set, and reporting confidence intervals for evaluation scores. This corresponds only to conservative significance testing and hides the variance induced by metaparameter variation. Time constraints prohibited the use of more sophisticated techniques for significance testing and variance component analysis.

\acks{
This work was partially funded by the Helmholtz Information \& Data Science School for Health (HIDSS4Health) and the German Research Foundation (DFG) through  Germany’s Excellence Strategy EXC 2181/1 – 390900948 (STRUCTURES).
}

\bibliography{ref}

\newpage
\appendix

\section{Features for Prediction Task}
\label{appendix:features}

\begin{table}[H]
\centering
\caption{Feature list for MIMIC-III: Besides the following 131 dynamic variables, only age and gender were extracted. The 15 variables marked with an asterisk are directly used for calculating the SOFA score.}
\label{tab:features_mimic}
\begin{tabular}{llll}

\toprule
ALP                  & Epinephrine*          & LDH                       & Packed RBC           \\
ALT                  & Famotidine            & Lactate                   & Pantoprazole         \\
AST                  & Fentanyl              & Lactated Ringers          & Phosphate            \\
Albumin              & FiO2*                 & Levofloxacin              & Piggyback            \\
Albumin 25\%         & Fiber                 & Lorazepam                 & Piperacillin         \\
Albumin 5\%          & Free Water            & Lymphocytes               & Platelet Count*      \\
Amiodarone           & Fresh Frozen Plasma   & Lymphocytes (Absolute)    & Potassium            \\
Anion Gap            & Furosemide            & MBP                       & Pre-admission Intake \\
BUN                  & GCS\_eye*             & MCH                       & Pre-admission Output \\
Base Excess          & GCS\_motor*           & MCHC                      & Propofol             \\
Basophils            & GCS\_verbal*          & MCV                       & RBC                  \\
Bicarbonate          & GT Flush              & Magnesium                 & RDW                  \\
Bilirubin (Direct)   & Gastric               & Magnesium Sulfate (Bolus) & RR                   \\
Bilirubin (Indirect) & Gastric Meds          & Magnesium Sulphate        & Residual             \\
Bilirubin (Total)*   & Glucose (Blood)       & Mechanically ventilated   & SBP*                 \\
CRR                  & Glucose (Serum)       & Metoprolol                & SG Urine             \\
Calcium Free         & Glucose (Whole Blood) & Midazolam                 & Sodium               \\
Calcium Gluconate    & HR                    & Milrinone                 & Solution             \\
Calcium Total        & Half Normal Saline    & Monocytes                 & Sterile Water        \\
Cefazolin            & Hct                   & Morphine Sulfate          & Stool                \\
Chest Tube           & Heparin               & Neosynephrine             & TPN                  \\
Chloride             & Hgb                   & Neutrophils               & Temperature          \\
Colloid              & Hydralazine           & Nitroglycerine            & Total CO2            \\
Creatinine Blood*    & Hydromorphone         & Nitroprusside             & Ultrafiltrate        \\
Creatinine Urine     & INR                   & Norepinephrine*           & Urine*               \\
D5W                  & Insulin Humalog       & Normal Saline             & Vancomycin           \\
DBP*                 & Insulin NPH           & O2 Saturation             & Vasopressin          \\
Dextrose Other       & Insulin Regular       & OR/PACU Crystalloid       & WBC                  \\
Dobutamine*          & Insulin largine       & PCO2                      & Weight               \\
Dopamine*            & Intubated             & PO intake                 & pH Blood             \\
EBL                  & Jackson-Pratt         & PO2*                      & pH Urine             \\
Emesis               & KCl                   & PT                        &                      \\
Eoisinophils         & KCl (Bolus)           & PTT                       &          \\
\bottomrule
\end{tabular}
\end{table}

\begin{table}[H]
\centering
\caption{Feature list for eICU: Besides the following 98 dynamic variables, there are 17 static variables covering age, gender, admission information, and ICU type. The 15 variables marked with an asterisk are directly used for calculating the SOFA score. \\
On the right column, there are 35 drug-related variables. Some of them seem redundant due to different hospitals but can not be merged because of different or not standardized concentrations.}
\label{tab:features_eicu}
\begin{tabular}{ll|l}
\toprule
ALP                & Lactate         & Amiodarone           \\
ALT                & Lymphocytes     & Dobutamine dose      \\
AST                & MBP             & Dobutamine ratio*    \\
Albumin            & MCH             & Dopamine dose        \\
Anion Gap          & MCHC            & Dopamine ratio*      \\
BUN                & MCV             & Epinephrine dose     \\
Base Deficit       & MPV             & Epinephrine ratio*   \\
Base Excess        & Magnesium       & Fentanyl 1           \\
Basophils          & Monocytes       & Fentanyl 2           \\
Bedside Glucose    & Neutrophils     & Fentanyl 3           \\
Bicarbonate        & O2 L/\%         & Furosemide           \\
Bilirubin (Direct) & O2 Saturation   & Heparin 1            \\
Bilirubin (Total)* & PT              & Heparin 2            \\
Bodyweight (kg)    & PTT             & Heparin 3            \\
CO2 (Total)        & PaCO2           & Heparin vol          \\
Calcium            & PaO2*           & Insulin 1            \\
Chloride           & Phosphate       & Insulin 2            \\
Creatinine (Blood)*& Platelets*      & Insulin 3            \\
Creatinine (Urine) & Potassium       & Midazolam 1          \\
DBP*               & Protein (Total) & Midazolam 2          \\
Eoisinophils       & RBC             & Milrinone 1          \\
EtCO2              & RDW             & Milrinone 2          \\
FiO2*              & RR              & Nitroglycerin 1      \\
Fibrinogen         & SBP*            & Nitroglycerin 2      \\
GCS eye*           & Sodium          & Nitroprusside        \\
GCS motor*         & Stool           & Norepinephrine 1     \\
GCS verbal*        & Temperature     & Norepinephrine 2     \\
Glucose            & Troponin - I    & Norepinephrine ratio*\\
HR                 & Urine*          & Pantoprazole         \\
Hct                & WBC             & Propofol 1           \\
Hgb                & pH              & Propofol 2           \\
INR                &                 & Propofol 3           \\
                   &                 & Vasopressin 1        \\
                   &                 & Vasopressin 2        \\
                   &                 & Vasopressin 3              \\
\bottomrule
\end{tabular}
\end{table}

\section{SOFA score}
\label{appendix:sofa}

The Sepsis-related Organ Failure Assessment (SOFA) is calculated by summing six subscores ranging from 0 to 4. In our setting, we had to recalculate MAP (mean arterial pressure) by SBP and DBP (systolic and diastolic blood pressure), the Horowitz coefficient PaO2/FiO2 by PaO2 and FiO2, but ignored the kind of mechanical ventilation. If no value for calculation in a SOFA subsystem was available, we took a value of 0.

\begin{table}[H]
	\centering
	\small
	\caption{SOFA score \citep{VincentETAL:96}. Abbreviations: CNS = Central nervous system; GCS = Glasgow Coma Scale; MV = mechanically ventilated including CPAP; MAP = mean arterial pressure, UO = Urine output.}
\begin{tabular}{rp{1.3cm}p{5cm}p{2.5cm}p{1.55cm}p{1.5cm}p{1.9cm}}
    \toprule
          & CNS         & Cardiovascular                                                                                                                        & Respiratory          & Coagu\-lation              & Liver             & Renal                                               \\
          \cmidrule{2-7}
    \rotatebox{90}{Score\hspace*{-1em}} & GCS         & MAP  \newline or vasopressors                                                                      & PaO2/FiO2 \newline(mmHg)       & Platelets \newline (×$10^3$/$\mu$l) & Bilirubin \newline(mg/dl) & Creatinine \newline(mg/dl)  or UO             \\ \cmidrule{2-7}
    +0    & 15          & MAP $\geq$ 70 mmHg                                                                                                                    & $\geq$ 400           & $\geq$ 150                 & $<$ 1.2     & $<$ 1.2                                       \\
    +1    & 13–14       & MAP $<$ 70 mmHg                                                                                                                 & $<$ 400        & $<$ 150              & 1.2–1.9           & 1.2–1.9                                             \\
    +2    & 10–12       & dopamine $\leq$ 5 $\mu$g/kg/min OR\newline dobutamine (any dose)                                                                      & $<$ 300        & $<$ 100              & 2.0–5.9           & 2.0–3.4                                             \\
    +3    & 6–9         & dopamine $>$ 5 $\mu$g/kg/min OR\newline epinephrine $\leq$ 0.1 $\mu$g/kg/min OR\newline norepinephrine $\leq$ 0.1 $\mu$g/kg/min              & $<$ 200 AND MV & $<$ 50               & 6.0–11.9          & 3.5–4.9 OR\newline $<$ 500 ml/day                   \\
    +4    & $<$ 6 & dopamine $>$ 15 $\mu$g/kg/min OR \newline epinephrine $>$ 0.1 $\mu$g/kg/min OR\newline norepinephrine $>$ 0.1 $\mu$g/kg/min & $<$ 100 AND MV & $<$ 20               & $>$ 12.0 & $>$ 5.0 OR \newline $<$ 200 ml/day\\\bottomrule
\end{tabular}
\end{table}

\section{SAPS-II score}
\label{appendix:sapsii}

\begin{landscape}
\begin{table}[H]
	\begin{threeparttable}[b]
	\centering
	\small
	\caption{List of  non-static components of SAPS-II score \citep{LeGallETAL:93}. 
			The total score is computed by summing the individual contributions. 
			Abbreviations: HR = heart rate; SBP = systolic blood pressure;  GCS = Glasgow Coma Scale; BUN = blood urea nitrogen; WBC = white blood cell count, MV = mechanically ventilated including CPAP.}
	\begin{tabular}{lcccccccccccc}
		\toprule
		Score & HR\tnote{1} & SBP\tnote{1} &  Temp\tnote{2}  & GCS\tnote{3} & PaO2/Fi02\tnote{3} & BUN\tnote{2} & Urine\tnote{4} & Sodium\tnote{1} & Potassium\tnote{1} & Bicarbonate\tnote{3} & WBC\tnote{1} & Bilirubin\tnote{2} \\
		      &    [1/s]    &    [mmHG]    & [\textdegree C] &              &     [mmHg/\%]      &   [mg/dl]    &    [ml/24h]    &    [mmol/L]     &      [mEq/L]       &       [mEq/L]        &  [1000/mm³]  &      [mg/dL]       \\ \midrule
		+0    &   70-119    &   100-199    &      else       &    14-15     &       not MV       &     $<$28      &   $\geq$1000   &     125-144     &      3.0-4.9       &       $\geq$20       &   1.0-19.9   &       $<$4.0       \\
		+1    &             &              &                 &              &                    &              &                &    $\geq$145    &                    &                      &              &                    \\
		+2    &    40-69    &    $>$199    &                 &              &                    &              &                &                 &                    &                      &              &                    \\
		+3    &             &              &    $\geq$39     &              &                    &              &                &                 &                    &        15-19         &   $\geq$20   &                    \\
		+4    &   120-159   &              &                 &              &                    &              &    500-599     &                 &                    &                      &              &      4.0-5.9       \\
		+5    &             &    70-99     &                 &    11-13     &                    &              &                &     $<$125      &  $<$3 or $\geq$5   &                      &              &                    \\
		+6    &             &              &                 &              &       $>$199       &    28-83     &                &                 &                    &        $<$15         &              &                    \\
		+7    &   $>$159    &              &                 &     9-10     &                    &              &                &                 &                    &                      &              &                    \\
		+9    &             &              &                 &              &      100-199       &              &                &                 &                    &                      &              &      $\geq$6       \\
		+10   &             &              &                 &              &                    &    $>$83     &                &                 &                    &                      &              &                    \\
		+11   &    $<$40    &              &                 &              &       $<$100       &              &     $<$500     &                 &                    &                      &              &                    \\
		+12   &             &              &                 &              &                    &              &                &                 &                    &                      &    $<$1.0    &                    \\
		+13   &             &    $<$70     &                 &     6-8      &                    &              &                &                 &                    &                      &              &                    \\
		+26   &             &              &                 &     $<$6     &                    &              &                &                 &                    &                      &              &                    \\ \bottomrule
		      &             &              &                 &              &
	\end{tabular}
	\begin{tablenotes}
		\item [1] worst (min or max) value in 24h
		\item [2] largest (max) value in 24h
		\item [3] smallest (min) value in 24h
		\item [4] total (sum) value in 24h
	\end{tablenotes}	
\end{threeparttable}
\end{table}
\end{landscape}

\section{Dataset densities}
\label{appendix:histograms}

\begin{figure}[H]
    \centering
    \includegraphics[width=0.75\linewidth]{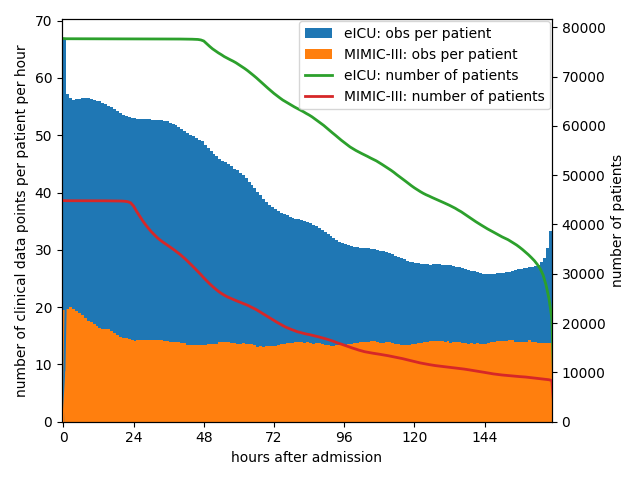}
    \caption{Comparison of the length of stay of patients (lines) and observations per hour per patient (bars) for MIMIC-III and eICU.}
    \label{fig:hist_occurence}
\end{figure}

\section{Metaparameter settings}
\label{appendix:metaparams}

\begin{table}[H]
\caption{Metaparameter settings for training runs on MIMIC-III. Best settings chosen on development data are shown in bold face.}
\begin{tabular}{@{}lll@{}}
\toprule
 & \multicolumn{2}{l}{MIMIC-III} \\ \midrule

 Parameter & Triplet & Dense \\
    \cmidrule(r){1-3}
    Embedding Size & 25, \textbf{50}, 100 & 128, 256, \textbf{512} \\
    Hidden Size Encoder & 25, \textbf{50}, 100 & 128, 256, \textbf{512} \\
    Hidden Size DMS Decoder & \textbf{50}, 100, 200 & 128, 256, \textbf{512} \\
    Hidden Size IMS Decoder & Output Dimensionality & Output Dimensionality \\
    \# Encoder Layers & 1, \textbf{2} & 1, \textbf{2} \\
    \# DMS Decoder Layers & \textbf{1}, 2 & \textbf{1}, 2 \\
    \# IMS Decoder Layers & \textbf{1}, 2 & \textbf{1}, 2 \\
    Learning Rate & 0.0001, \textbf{0.0005}, 0.001 & 0.0001, \textbf{0.0005}, 0.001 \\
    Batch Size & 8, 16, \textbf{32} & 8, 16, \textbf{32} \\
    Attention Heads Encoder & 2, \textbf{4}, 8 & 2, 4, \textbf{8} \\
    Attention Heads DMS Decoder & Prediction Timesteps & Prediction Timesteps \\
    Attention Heads IMS Decoder & \textbf{1}, 2, 4 & \textbf{1}, 2, 4 \\
    Dropout & 0.05, 0.1, \textbf{0.2} & \textbf{0.05}, 0.1, 0.2 \\
    Epochs & 100 & 100 \\
    Patience & 6 & 6 \\
    Random Seed & Unixtime variation & Unixtime variation\\
\bottomrule
\end{tabular}
\end{table}

\begin{table}[H]
\caption{Metaparameter settings for training runs on eICU. Best settings chosen on development data are shown in bold face.}
\begin{tabular}{@{}lll@{}}
\toprule
 & \multicolumn{2}{l}{eICU} \\ \midrule
Parameter & Triplet & Dense \\
    \cmidrule(r){1-3}
    Embedding Size & 25, \textbf{50}, 100 & \textbf{256}, 512, 1024 \\
    Hidden Size Encoder & 25, \textbf{50}, 100 & \textbf{256}, 512, 1024 \\
    Hidden Size DMS Decoder & 25, \textbf{50}, 100 & \textbf{256}, 512, 1024 \\
    Hidden Size IMS Decoder & Output Dimensionality & Output Dimensionality \\
    \# Encoder Layers & 1, \textbf{2}, 3 & 1, \textbf{2}, 3 \\
    \# DMS Decoder Layers & \textbf{1} & \textbf{1} \\
    \# IMS Decoder Layers & \textbf{1} & \textbf{1} \\
    Learning Rate & \textbf{0.0005} & \textbf{0.0005} \\
    Batch Size & \textbf{32} & \textbf{32} \\
    Attention Heads Encoder & 2, \textbf{4}, 8 & \textbf{8} \\
    Attention Heads DMS Decoder & Prediction Timesteps & Prediction Timesteps \\
    Attention Heads IMS Decoder & \textbf{1},2,4 & \textbf{1},2,4 \\
    Dropout & \textbf{0.2} & \textbf{0.05} \\
    Epochs & 600 & 600 \\
    Patience & 6 & 6 \\
    Random Seed & Unixtime variation & Unixtime variation\\
\bottomrule
\end{tabular}
\end{table}

\section{Experiments on eICU}
\label{appendix:eicu}

\begin{table}[H]
	\centering
	\small
	\caption{Main evaluation results on eICU for combinations of sparse and dense encoders with direct multi-step (DMS) and iterative multi-step (IMS) decoders, trained by teacher forcing (TF), student forcing (SF), or scheduled sampling (SS), with or without backpropagation (BP) through predictions. Only curricula for randomized increasing (RI) scheduling are shown. Evaluation is done according to MSE of all variables (Equation \ref{eq:mse}), MSE for SOFA (Equation \ref{eq:SOFA-MSE}), and Accuracy for Sepsis (Equation \ref{eq:sepsis_acc}). Numbers in subscripts denote the 95\% confidence interval for the estimation of the respective evaluation score on the test set. Best results are shown in bold face.}
	\label{tab:eicu_main_results}
	\begin{tabular}{lllllccc}
		\toprule
  Model      & Enc & Dec & Train& BP  &            MSE             &          MSE-SOFA          &                 Acc-Sepsis                  \\ \midrule
1&Triplet&DMS&-&-&5.748$_{[5.747,5.749]}$&2.673$_{[2.672,2.673]}$&87.59$_{[86.54,88.64]}$\\
2&Triplet&IMS&TF&-&9.175$_{[9.174,9.176]}$&4.429$_{[4.429,4.429]}$&83.97$_{[82.80,85.14]}$\\
3&Triplet&IMS&SF&No&\textbf{5.134}$_{[5.133,5.135]}$&2.113$_{[2.113,2.114]}$&87.05$_{[86.09,88.00]}$\\
4&Triplet&IMS&SF&Yes&5.160$_{[5.159,5.161]}$&2.295$_{[2.295,2.295]}$&87.40$_{[86.42,88.38]}$\\
9&Triplet&IMS&SS-RI&No&5.256$_{[5.255,5.257]}$&2.233$_{[2.232,2.233]}$&86.98$_{[86.02,87.94]}$\\
10&Triplet&IMS&SS-RI&Yes&5.331$_{[5.330,5.332]}$&2.044$_{[2.243,2.244]}$&86.27$_{[85.28,87.26]}$\\
\midrule
13&Dense&DMS&-&-&5.352$_{[5.351,5.353]}$&2.321$_{[2.321,2.321]}$&87.88$_{[87.18,88.58]}$\\
14&Dense&IMS&TF&-&9.104$_{[9.102,9.105]}$&4.292$_{[4.291,4.292]}$&84.92$_{[83.92,85.91]}$\\
15&Dense&IMS&SF&No&5.528$_{[5.527,5.529]}$&2.125$_{[2.125,2.126]}$&90.02$_{[89.07,90.98]}$\\
16&Dense&IMS&SF&Yes&5.395$_{[5.394,5.396]}$&\textbf{2.024}$_{[2.024,2.024]}$&\textbf{90.66}$_{[89.73,91.58]}$\\
21&Dense&IMS&SS-RI&No&5.322$_{[5.321,5.323]}$&2.273$_{[2.273,2.274]}$&88.34$_{[87.45,89.24]}$\\
22&Dense&IMS&SS-RI&Yes&5.399$_{[5.398,5.300]}$&2.208$_{[2.208,2.209]}$&88.95$_{[88.03,89.86]}$\\
\midrule
Informer&Dense&DMS&-&-&5.578$_{[5.677,5.679]}$&2.305$_{[2.304,2.305]}$&87.99$_{[87.29,88.68]}$\\
Autoformer&Dense&DMS&-&-&5.984$_{[5.983,5.986]}$&2.254$_{[2.254,2.255]}$&85.21$_{[84.22,86.19]}$\\
DLinear&Dense&DMS&-&-&5.974$_{[5.973,5.975]}$&3.340$_{[3.340,3.340]}$&85.10$_{[84.11,86.09]}$\\
Linear&Dense&DMS&-&-&5.975$_{[5.974,5.976]}$&3.331$_{[3.330,3.331]}$&85.07$_{[84.08,86.07]}$\\
\bottomrule
	\end{tabular}
\end{table}

\begin{table}[H]
	\centering
	\small
	\caption{SAPS-II evaluation results on eICU for combinations of sparse and dense encoders with direct multi-step (DMS) and iterative multi-step (IMS) decoders, trained by teacher forcing (TF), student forcing (SF), or scheduled sampling (SS), with or without backpropagation (BP) through predictions. Only curricula for randomized increasing (RI) scheduling are shown. Evaluation is done according to MSE of all variables (Equation \ref{eq:mse}), MSE for SOFA (Equation \ref{eq:SOFA-MSE}), and Accuracy for Sepsis (Equation \ref{eq:sepsis_acc}). Numbers in subscripts denote the 95\% confidence interval for the estimation of the respective evaluation score on the test set. Best results are shown in bold face.}
	\label{tab:eicu_saps_results}
	\begin{tabular}{lllllcc}
		\toprule
  Model      & Enc & Dec & Train& BP  &            MSE             &          MSE-SAPS-II           \\ \midrule
1&Triplet&DMS&-&-&5.748$_{[5.747,5.749]}$&93.799$_{[93.799,93.799]}$\\
2&Triplet&IMS&TF&-&9.175$_{[9.174,9.176]}$&123.362$_{[123.362,123.363]}$\\
3&Triplet&IMS&SF&No&\textbf{5.134}$_{[5.133,5.135]}$&91.337$_{[91.337,91.337]}$\\
4&Triplet&IMS&SF&Yes&5.160$_{[5.159,5.161]}$&94.379$_{[94.379,94.380]}$\\
9&Triplet&IMS&SS-RI&No&5.256$_{[5.255,5.257]}$&96.613$_{[96.613,96.614]}$\\
10&Triplet&IMS&SS-RI&Yes&5.331$_{[5.330,5.332]}$&91.571$_{[91.571,91.571]}$\\
\midrule
13&Dense&DMS&-&-&5.352$_{[5.351,5.353]}$&89.429$_{[89.429,89.429]}$\\
14&Dense&IMS&TF&-&9.104$_{[9.102,9.105]}$&121.478$_{[121.478,121.479]}$\\
15&Dense&IMS&SF&No&5.528$_{[5.527,5.529]}$&87.598$_{[87.598,87.599]}$\\
16&Dense&IMS&SF&Yes&5.395$_{[5.394,5.396]}$&\textbf{86.279}$_{[86.279,86.280]}$\\
21&Dense&IMS&SS-RI&No&5.322$_{[5.321,5.323]}$&92.525$_{[92.525,92.526]}$\\
22&Dense&IMS&SS-RI&Yes&5.399$_{[5.398,5.300]}$&91.827$_{[91.827,91.828]}$\\
\midrule
Informer&Dense&DMS&-&-&5.578$_{[5.577,5.579]}$&89.139$_{[89.138,89.139]}$\\
Autoformer&Dense&DMS&-&-&5.984$_{[5.983,5.985]}$&89.513$_{[89.513,89.514]}$\\
DLinear&Dense&DMS&-&-&5.974$_{[5.973,5.975]}$&94.177$_{[94.176,94.177]}$\\
Linear&Dense&DMS&-&-&5.975$_{[5.974,5.976]}$&93.743$_{[93.743,93.743]}$\\
\bottomrule
	\end{tabular}
\end{table}

\section{Detailed Results for Sepsis Prediction}
\label{appendix:f1}

\begin{table}[H]
	\centering
	\small
	\caption{F1 score and the values of the confusion matrix (TP, FP, FN, TN) used to compute it for the MIMIC-III dataset.}
	\label{tab:mimic_f1_table}
	\begin{tabular}{lllllccccc}
		\toprule
Model&Enc&Dec&Train&BP&TP&FP&FN&TN&F1\\ \midrule
1&Triplet&DMS&-&- &84&41&41&662&67.2\\
2&Triplet&IMS&TF&- &84&165&41&538&44.92\\
3&Triplet&IMS&SF&No &84&46&41&657&65.88\\
4&Triplet&IMS&SF&Yes &84&48&41&655&65.37\\
9&Triplet&IMS&SS-RI&No &84&78&41&625&58.54\\
10&Triplet&IMS&SS-RI&Yes &84&71&41&632&60.0\\
13&Dense&DMS&-&- &84&76&41&627&58.95\\
14&Dense&IMS&TF&- &81&236&44&467&36.65\\
15&Dense&IMS&SF&No &80&46&45&656&63.75\\
16&Dense&IMS&SF&Yes &85&40&40&663&68.0\\
21&DENSE&IMS&SS-RI&No &85&72&40&631&60.28\\
22&DENSE&IMS&SS-RI&Yes &85&82&40&621&58.22\\
Informer&Dense&DMS&-&- &77&70&48&633&56.62\\
Autoformer&Dense&DMS&-&- &89&50&36&653&67.42\\
DLinear&Dense&DMS&-&- &77&149&48&554&43.87\\
Linear&Dense&DMS&-&- &77&140&48&563&45.03\\
\bottomrule
	\end{tabular}
\end{table}

\begin{table}[H]
	\centering
	\small
	\caption{F1 score and the values of the confusion matrix (TP, FP, FN, TN) used to compute it for the eICU dataset.}
	\label{tab:eicu_f1_table}
	\begin{tabular}{lllllccccc}
		\toprule
Model&Enc&Dec&Train&BP&TP&FP&FN&TN&F1\\ \midrule
1&Triplet&DMS&-&- &83&181&289&3236&26.1\\
2&Triplet&IMS&TF&- &46&281&326&3136&13.16\\
3&Triplet&IMS&SF&No &71&190&301&3227&22.43\\
4&Triplet&IMS&SF&Yes &75&180&297&3237&23.92\\
9&Triplet&IMS&SS-RI&No &71&192&301&3215&22.36\\
10&Triplet&IMS&SS-RI&Yes &76&194&326&3223&22.62\\
13&Dense&DMS&-&- &84&171&288&3246&26.79\\
14&Dense&IMS&TF&- &69&269&303&3148&19.44\\
15&Dense&IMS&SF&No &98&104&274&3313&34.15\\
16&Dense&IMS&SF&Yes &90&72&282&3345&33.71\\
21&Dense&IMS&SS-RI&No &75&145&297&3272&25.34\\
22&Dense&IMS&SS-RI&Yes &86&132&286&3285&29.15\\
Informer&Dense&DMS&-&- &84&167&288&3250&26.97\\
Autoformer&Dense&DMS&-&- &72&260&300&3157&20.45\\
DLinear&Dense&DMS&-&- &57&249&315&3168&16.81\\
Linear&Dense&DMS&-&- &57&250&315&3167&16.79\\
\bottomrule
	\end{tabular}
\end{table}

\end{document}